\documentclass[]{article} % For LaTeX2e
\usepackage{iclr2025_conference,times}
\iclrfinalcopy
\usepackage{amsmath,amsfonts,bm}

\def\eqref#1{equation~\ref{#1}}
\def\1{\bm{1}}

\DeclareMathAlphabet{\mathsfit}{\encodingdefault}{\sfdefault}{m}{sl}
\SetMathAlphabet{\mathsfit}{bold}{\encodingdefault}{\sfdefault}{bx}{n}

\usepackage{hyperref}
\usepackage{url}
\usepackage{graphicx}
\usepackage{subfig}
\usepackage{algorithm}
\usepackage{algpseudocode}
\usepackage{multirow}
\usepackage{makecell}
\usepackage{longtable}
\usepackage{booktabs}

\title{LogQuant: Log-Distributed 2-Bit Quantization of KV Cache with Superior Accuracy Preservation}

\author{Han Chen \& Zining Zhang \& Bingsheng He \\
School of Computing\\
National University of Singapore\\
21 Lower Kent Ridge Road, Singapore 119077 \\
\texttt{\{chenhan, zzn\}@u.nus.edu, hebs@comp.nus.edu.sg} \\
\And
Zicong Jiang \\
School of Electronic and Information Engineering \\
South China University of Technology \\
381 Wushan Road, Tianhe District, Guangzhou, 510641 P. R. China \\
\texttt{202420111170@mail.scut.edu.cn} \\
\AND
Pingyi Luo \& Mian Lu \& Yuqiang Chen \\
4Paradigm \\
\#03-20 Galaxis (West Lobby),Singapore 138522 \\
\texttt{\{luopingyi, lumian, chenyuqiang\}@4paradigm.com}
}

\begin{document}

\maketitle

\begin{abstract}

We introduce LogQuant, a groundbreaking 2-bit quantization technique for KV Cache in large language model (LLM) inference, delivering substantial memory savings while preserving superior performance. Previous methods either assume that later tokens are more important or attempt to predict important tokens based on earlier attention patterns. Both approaches, however, can result in performance bottlenecks or frequent mispredictions.

LogQuant takes a different approach. By applying a log-based filtering mechanism, it selectively compresses the KV Cache across the entire context, achieving better performance with the same or even reduced memory footprint compared to existing methods. In benchmark tests, it enhances throughput by 25\% and boosts batch size by 60\% without increasing memory consumption. For challenging tasks such as Math and Code Completion, LogQuant improves accuracy by 40\% to 200\% at the same compression ratio, outperforming comparable techniques. LogQuant integrates effortlessly with popular inference frameworks like Python’s \texttt{transformers} library. %and will be made open-source upon publication.
Implementation can be available in \href{https://github.com/Concyclics/LogQuantKV}{https://github.com/Concyclics/LogQuantKV}.

\end{abstract}

\section{Introduction}
\label{sec:intro}
The rapid evolution of Large Language Models (LLMs) has enabled context window expansion from 4k to 128k tokens~\citep{llama31, openaiapi}, driving demand for efficient KV cache management in applications like multi-round {chatbot conversations}~\citep{openaiapi, claudechat, DeepSeekchat} and {document-based question answering}~\citep{gao2023retrieval, lewis2020retrieval}, where comprehensive contextual understanding is required. Moreover, reasoning models such as OpenAI o1~\citep{openaio1}, increased the demand for even longer reasoning contexts, xacerbated the memory challenges faced in KV cache management.

\begin{figure}[hbt]
    \centering
    \includegraphics[width=0.8\linewidth]{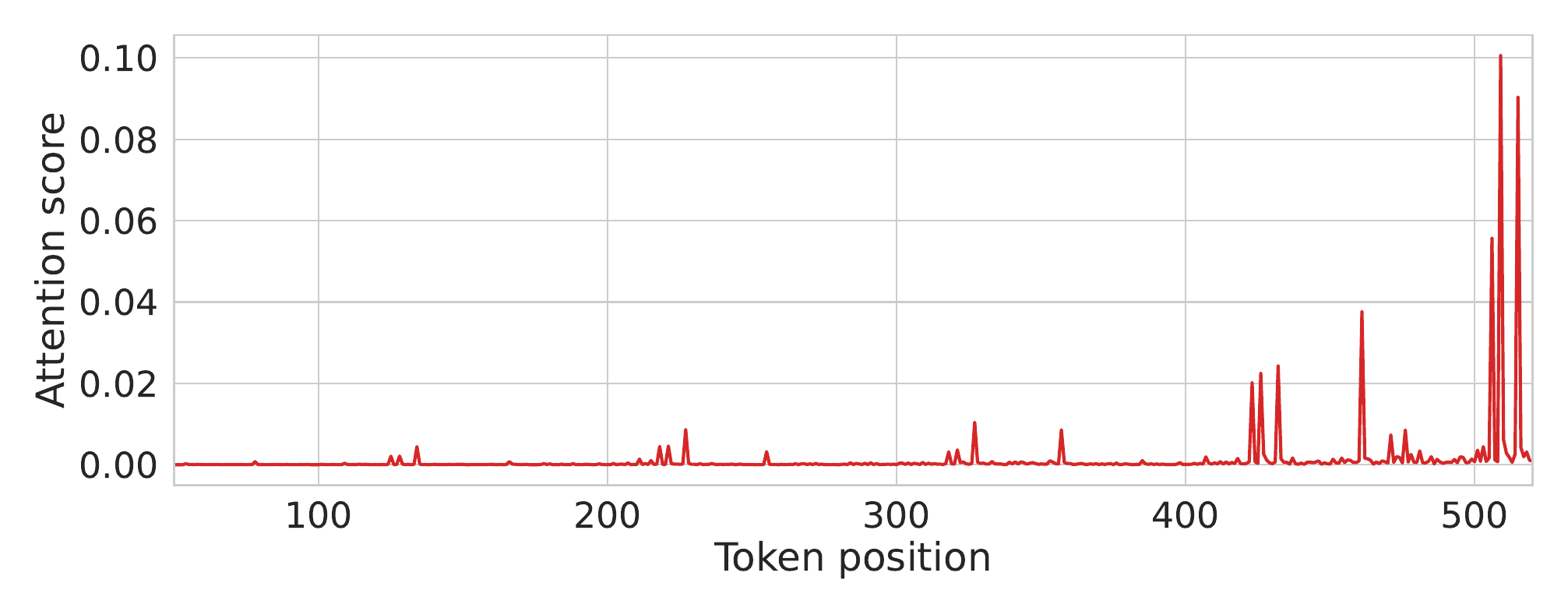}%
    \caption{The observed log-distribution pattern is evident not only in the magnitude of attention scores but also in the positions of attention spikes. These spikes become sparser as the model attends to tokens further from the most recent position, indicating that the model not only focuses on nearby tokens. This phenomenon, illustrated here with Llama3-8B-Instruct~\citep{dubey2024llama3} on the GSM8K dataset~\citep{GSM8K}, is consistent across different tasks and models, as further detailed in Section~\ref{sec:Methodology}.}
    \label{fig:log_position}
\end{figure}

Recent studies \cite{zhang2024h2o, li2024snapkv, dong2024qaq} reveal KV cache's linear memory growth with context length and even exceeds model weights in long context and batch inference, posing serious deployment challenges. Existing KV Cache compression methods adopt either \textit{eviction}, (H2O~\citep{zhang2024h2o}, Keyformer~\citep{adnan2024keyformer}, snapKV~\citep{li2024snapkv}), aim to reduce memory usage by selectively removing tokens deemed unimportant. or \textit{quantization} (QAQ~\citep{dong2024qaq}, KiVi~\citep{liu2024kivi}), reduce the precision of less important tokens, retaining more data while minimizing memory costs. Both struggle with importance identification. window-based methods (KiVi, StreamingLLM~\citep{streamingLLM}) risk missing distant important tokens, while attention-based approaches (H2O, keyformer) suffer prediction errors from historical scores.

Our approach addresses these shortcomings by leveraging a key insight: the positions of the \textit{attention spikes} (i.e. high attention scores) follow a log distribution as shown in Figure~\ref{fig:log_position}, resulting in sparser importance for tokens as they move further from the current position. By utilizing this property, we can outperform existing methods across a wide range of tasks. Additionally, the original absolute positions of KV cache entries can be disregarded without changing the final attention results during the decoding phase, which allows us to enhance the speed of our log-distributed quantization method.

The key contributions of this paper are as follows:
\begin{itemize}
    \item \textbf{Observation of Log-Distributed Attention Spikes}: We observe that in various models and downstream tasks, the positions of high attention spikes follow a log distribution, becoming sparser as tokens move further from the current position. This insight underpins our approach to estimate token importance.
    
    \item \textbf{Design of LogQuant}: Leveraging this log-distribution observation, we introduce LogQuant, a 2-bit quantization technique that significantly improves accuracy. LogQuant outperforms existing methods like KiVi and H2O by better preserving important tokens, achieving a 40\% to 200\% improvement in accuracy on complex tasks such as Math and Code Completion with the same or higher compression ratio.
    
    \item \textbf{Throughput Optimization}: By ignoring the absolute positions of KV cache entries, our method further optimizes the speed of quantization/dequantization process without affecting the final attention results, resulting in a 25\% increase in throughput and a 60\% increase in batch size.
\end{itemize}

The remainder of the paper is organized as follows: Section~\ref{sec:Methodology} details the core concepts behind our proposed LogQuant methods, Section~\ref{sec:Experiments} present an extensive set of experiments, Section~\ref{sec:Conclusion} summarizes our findings and discusses potential directions for future work.

\section{Methodology}
\label{sec:Methodology}

In Section~\ref{sec:Preliminary}, we analyze the distribution of attention scores and evaluate the impact of quantization loss, both with and without sink tokens. Section~\ref{sec:log_distributed_attention_pattern} explores the distribution of token importance and introduces our log-based selection strategy. In Section~\ref{sec:loss}, we compare the effects of quantization and eviction under this selection scheme, demonstrating the superiority of quantization over eviction. To further enhance efficiency, Section~\ref{sec:position_agnostic_attention_calculation} prove that attention computation is position-agnostic. Finally, we present the implementation details of our proposed \textbf{LogQuant} method in Section~\ref{sec:LogQuant}.

\subsection{Preliminary Study of KV Cache and Attention Scores}
\label{sec:Preliminary}
\begin{figure}[tb]
    \centering
    \includegraphics[width=1\linewidth]{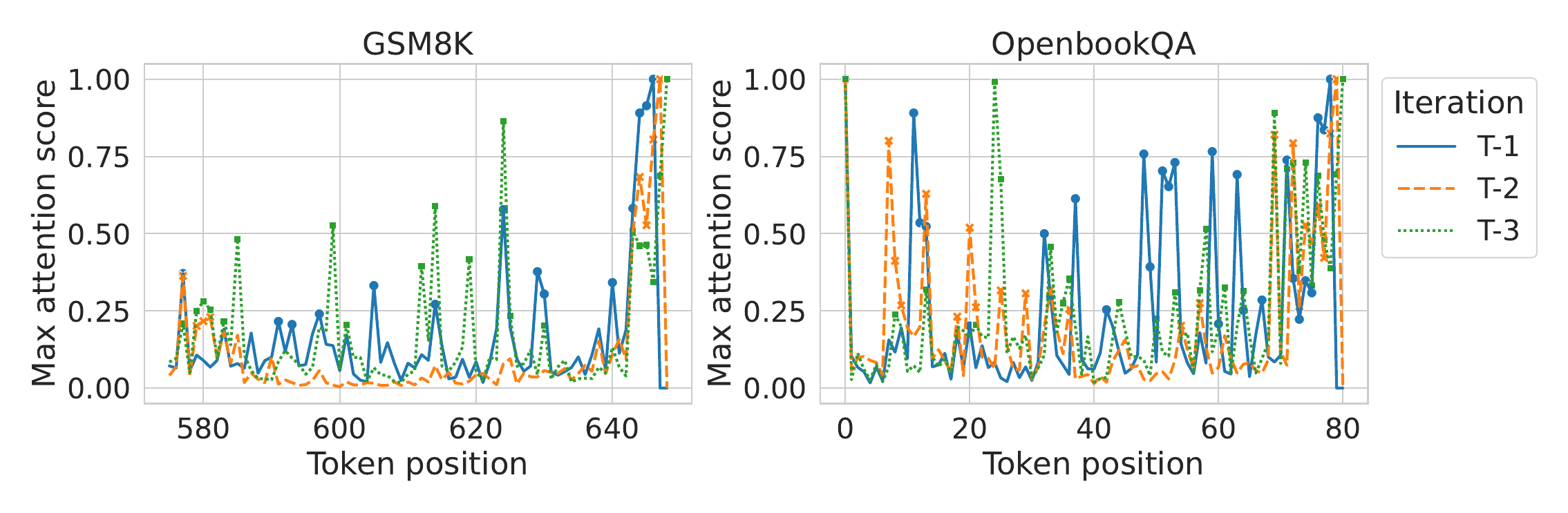}
    \caption{The maximum attention score of each token position across four consecutive decoding steps, marking the high attention positions for illustrating the unpredictable nature of attention scores. This analysis was conducted using Llama3-8B-Instruct~\citep{dubey2024llama3} on the GSM8K~\citep{GSM8K} and OpenBookQA~\citep{OpenBookQA2018} datasets.}
    \label{fig:dynamic}
\end{figure}

There are two well-established observations in recent works particularly relevant to KV cache compression. First, many tokens exhibit consistently low attention scores, indicating that their KV cache entries can be safely compressed with minimal impact on performance~\citep{liu2024kivi}. Second, predicting token importance based on previous decoding steps is unreliable, as attention scores can vary significantly across iterations, making it difficult to accurately identify which tokens should be preserved~\citep{dong2024qaq, jiang2024minference}. This is also demonstrated in Figure~\ref{fig:dynamic}.

\begin{figure}[ht]
    \centering
    \includegraphics[width=0.95\linewidth]{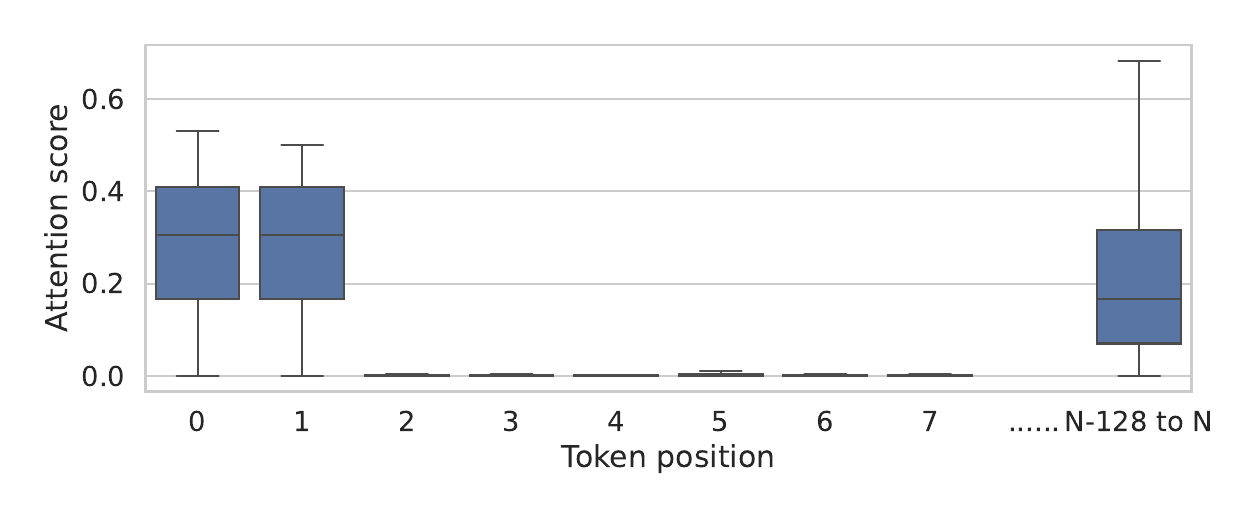}
    \caption{Attention distribution across different token positions, represented as boxplots based on 25\% quantiles across all attention heads. The median and overall distribution of attention scores for sink tokens~\citep{streamingLLM} (tokens 0 and 1) are greater than the sum of the most recent 128 tokens. The attention scores are derived from experiments using Llama3-8B-Instruct~\citep{dubey2024llama3} and the GSM8K~\citep{GSM8K} dataset.}
    \label{fig:position}
\end{figure}

Inspired by the observation of \textit{sink tokens}~\citep{streamingLLM}, which are the first few tokens that consistently receive high attention scores (Figure~\ref{fig:position}), we included these tokens in the set maintained at original precision to improve accuracy in 2-bit quantization. However, as shown in Table~\ref{tab:KiVi+Sink}, this adjustment yielded minimal improvement. This suggests that while sink tokens play a role in defining the conversational context, maintaining high precision for only these tokens is insufficient, indicating that tokens beyond the first few are also crucial for preserving model performance.

\begin{table}[ht]
\caption{\centering{Impact of retaining the first two tokens (referred to as "Sink") at original precision.} The final answer accuracy results on GSM8K~\cite{GSM8K} are presented. We present the improvement as $\Delta_{\text{Sink}}$. Both methods maintain the recent 128 tokens at original precision.}
\centering 
\resizebox{1.0\columnwidth}{!}{
\begin{tabular}{cccccr}
\hline
\textbf{Model}  & \textbf{baseline(BF16)} & \textbf{KiVi(4-bit)} & \textbf{KiVi(2-bit)} & \textbf{KiVi(2-bit)+Sink(BF16)} & \textbf{$\Delta_{Sink}$} \\ \hline
\\ 
Llama3.1-8B-Instruct  & 71.41 & 67.24  & 18.04   & 18.49  & +0.45    \\ 
Qwen1.5-7B-Chat & 57.24 & 52.27 & 39.80  & 39.42 & -0.38      \\ \hline
\end{tabular}
}
\label{tab:KiVi+Sink}
\end{table}

\subsection{The Log-distributed Attention Pattern}
\label{sec:log_distributed_attention_pattern}

As mentioned in Section~\ref{sec:intro}, our analysis of attention heads reveals a log-distributed high-attention pattern, which motivates the development of a quantization scheme that follows this distribution. We introduce a selection scheme where a window of size $2W$ retains the most recent consecutive tokens in full precision. Following this, another window of size $W/2$ selects tokens spaced one token apart, and then a window of size $W/4$ follows the similar pattern and so on. Finally, a window of $3W$ tokens is reserved in full precision. This creates a log-distributed token selection scheme.

\begin{figure}[tb]
    \centering
    % First subfigure
    \includegraphics[width=1\linewidth]{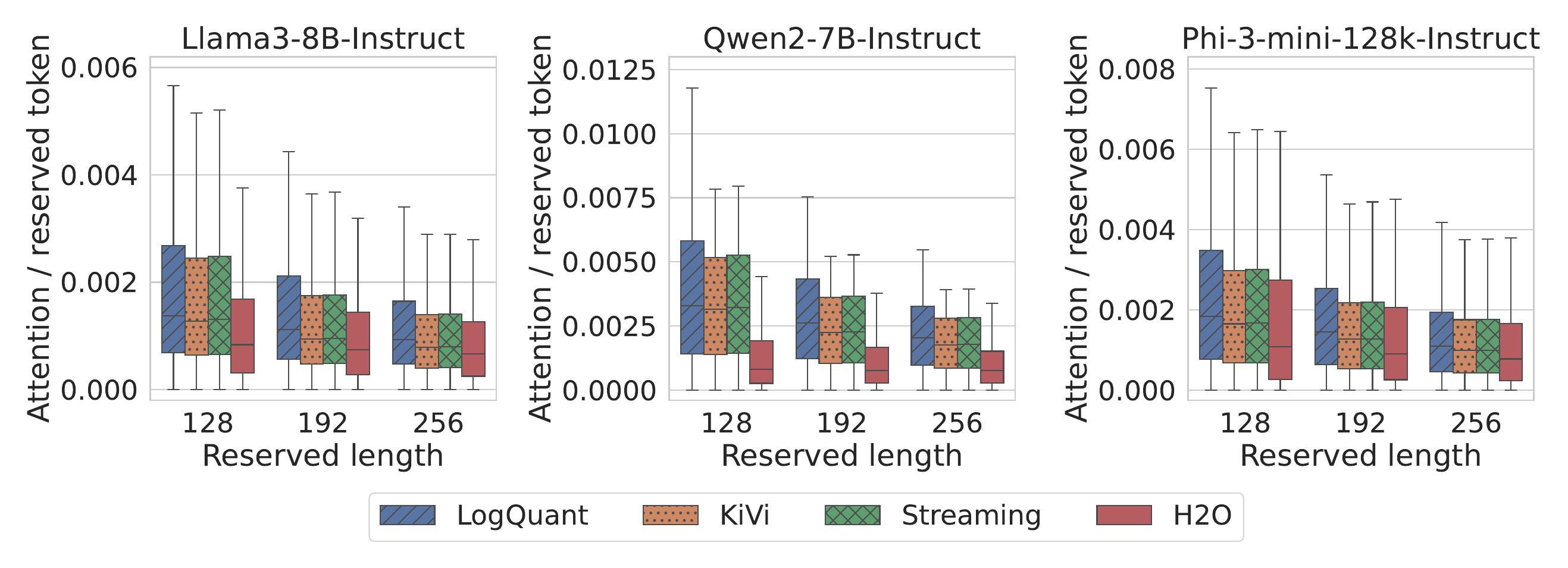}
    \caption{The attention coverage without the first two sink tokens for different {selection methods}~\citep{liu2024kivi, streamingLLM, zhang2024h2o} and different {models}~\citep{dubey2024llama3, yang2024qwen2, abdin2024phi}, tested on a subset of the GSM8K~\citep{GSM8K} dataset. Details of LogQuant will be introduced in Section \ref{sec:LogQuant}.}
    \label{fig:coverage}
\end{figure}

We compare this log-distributed selection to other methods: KiVi, which selects only the most recent $3W$ tokens; StreamingLLM, which selects the most recent $3W$ tokens plus the first four \textit{sink tokens}; and H2O, which uses previous attention scores to select the top $3W$ tokens. To evaluate these methods, we define \textit{token coverage} as the average attention score captured by the selection scheme: 
\begin{equation}
\begin{aligned}
\text{Token Coverage} = \frac{\sum_{i=1}^{3W}{\text{Attention Score of Selected Tokens}}}{3W}.\\
\end{aligned}
\end{equation}
Figure~\ref{fig:coverage} presents the results, where we exclude the first two tokens for calibration, as they typically have high attention scores but contribute minimally to overall model performance (see Section~\ref{sec:Preliminary}).

The results demonstrate that our log-distributed selection scheme covers high-attention tokens more effectively. This suggests that filtering tokens for quantization based on this log distribution leads to better token importance preservation.

\subsection{Comparison of Quantization and Eviction Strategies}
\label{sec:loss}

When implementing log-distributed token selection for KV Cache compression, two primary approaches emerge: quantization and eviction. These methods differ fundamentally in their operation. Quantization reduces the numerical precision of individual tokens, whereas eviction removes tokens entirely, thereby shortening the sequence length.

This distinction becomes critical due to the nature of the attention mechanism. The softmax function normalizes attention scores such that their sum equals 1. Consequently, removing tokens through eviction creates larger deviations from the original attention distribution compared to precision reduction via quantization. Specifically, eviction eliminates certain tokens from the attention computation entirely, while quantization retains all tokens with reduced numerical accuracy.

\begin{figure}
    \centering
    \includegraphics[width=1\linewidth]{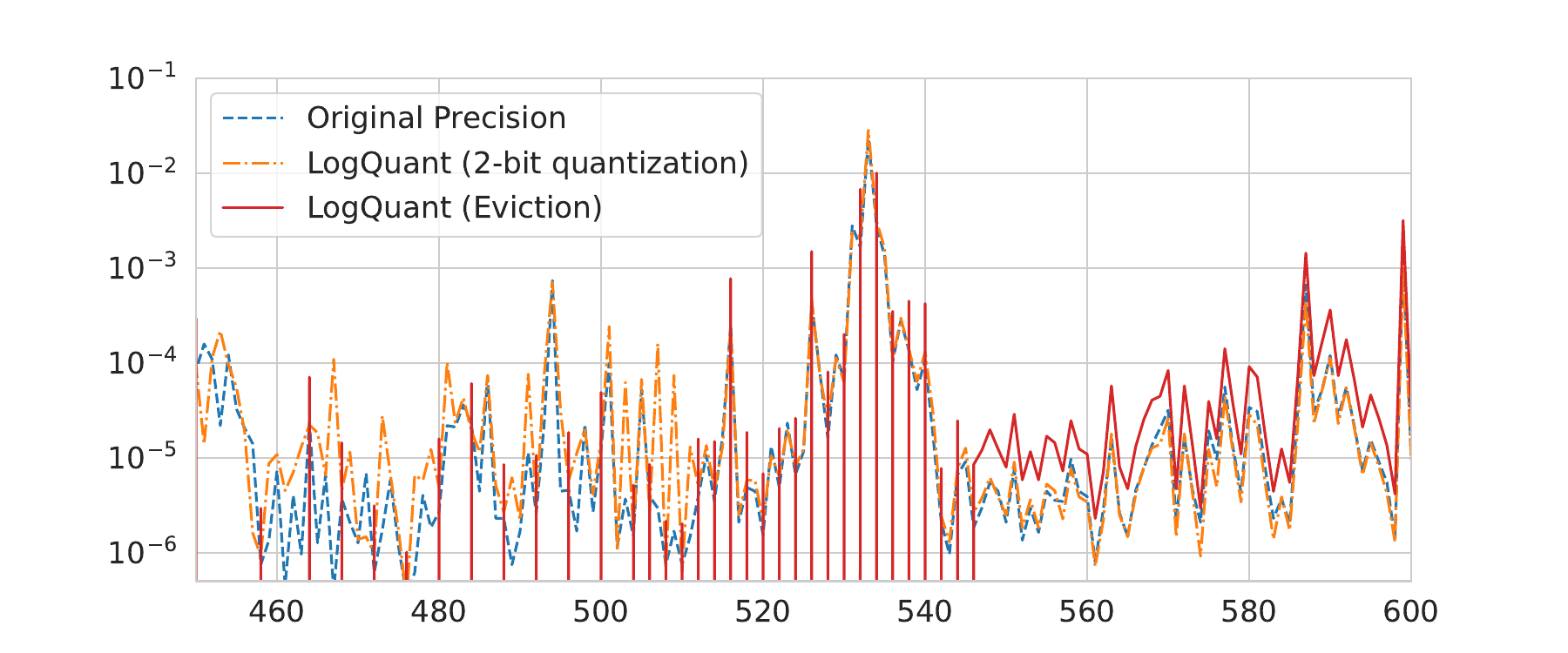}
    \caption{Eviction and Quantization Loss on Attention Distribution}
    \label{fig:eviction_loss}
\end{figure}

As demonstrated in \autoref{fig:eviction_loss}, this behavioral difference is visually apparent. Quantitative results on the GSM8K dataset using Llama3.1-8B (see \autoref{tab:l1_error_comparison}) show that eviction-based methods produce twice and higher attention errors than quantization. Based on these findings, we select quantization as the compression strategy.

\begin{table}[h!]
\centering
\caption{Comparison of L1 error with original attention for eviction and quantization.}
\label{tab:l1_error_comparison}

\begin{tabular}{cccc}
%\toprule
\hline
 \textbf{LogQuant (2-bit)} & \textbf{KiVi (2-bit)} & \textbf{LogQuant (Eviction)} & \textbf{KiVi (Eviction)} \\
%\midrule
\hline
\\
 432.50 & 556.10 & 1076.70 & 1612.56 \\
 \hline
%\bottomrule
\end{tabular}

\end{table}

\subsection{Position-Agnostic Attention Calculation}
\label{sec:position_agnostic_attention_calculation}
LLM inference involves two phases: prefill and decoding (Section~\ref{sec:RelatedWork}). As described in \cite{yuan2024llm_bottleneck}, the decoding phase is computationally expensive and memory-bound due to the use of the KV Cache. In the prefill phase, the model processes the input prompt in a single pass. However, during decoding, new tokens are generated one at a time, and each generation step requires access to the entire KV Cache. This leads to inefficiencies in both memory usage and execution time.

To mitigate these inefficiencies, we plan to accelerate the attention procedure. The attention operation can be expressed mathematically as follows:

\begin{equation}
\begin{aligned}
    A &= \text{Softmax}(Q \cdot K^T) \\
    O &= A \cdot V,
\end{aligned}
\end{equation}

where \( A \) is the attention distribution, a \( 1 \times N \) vector resulting from the softmax operation applied to the product of \( Q \) and the transpose of \( K \) and \( O \) is the output, a \( 1 \times d \) vector calculated by multiplying the attention distribution \( A \) with the Value matrix \( V \).

Since the attention distribution \( A \) aggregates values over all \( N \) tokens, the specific ordering of tokens in the Key and Value matrices does not affect the final output. This property allows us to permute or reorder the Key and Value caches without any loss of accuracy.
By leveraging this insight, we can optimize the KV Cache by concatenating high-precision tokens with quantized tokens while disregarding their original positions. This approach enhances memory locality and processing efficiency while maintaining the correctness of the attention computation.
This leads to the relation:

\begin{equation}
    A \cdot V = A_{P} \cdot V_{P},
\end{equation}

where \( P \) is a permutation of the indices \( \{1, \ldots, N\} \). This enables us to optimize the KV Cache effectively.

\subsection{LogQuant: Algorithm and Implementation}
\label{sec:LogQuant}
\begin{figure}[tb]
    \centering
    \includegraphics[width=0.9\linewidth]{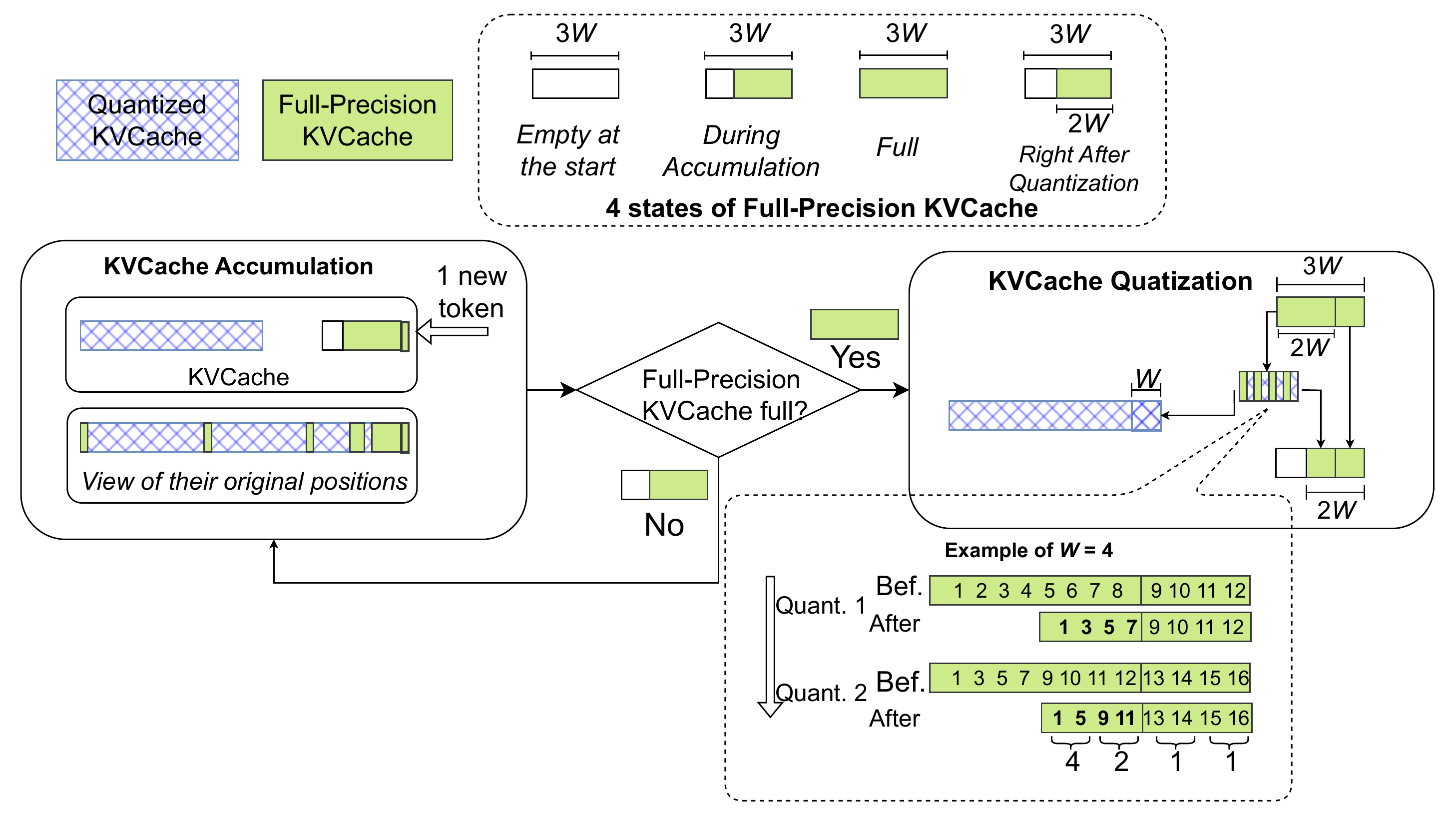}
    \caption{LogQuant's KV cache compression workflow. The number of reserved original-precision tokens increases from \(2W\) to \(3W\). We then apply a log-sparse strategy to filter the first \(2W\) tokens, quantize half of these tokens, and compress the reserved token length back to \(2W\).}
    \label{fig:workflow}
\end{figure}

\textbf{Algorithm.} After comparing different logarithmic bases \(\log_N\), we found that a base-2 logarithmic implementation is sufficiently effective for our purposes. To maintain logarithmic sparsity within a specified length, we adopt this base-2 logarithmic approach. 
We fix a window length configuration \(W\), allowing us to retain up to \(3W\) tokens at original precision. Each time the length limit is reached, we reduce the density of tokens in the first two windows (each of length \(W\)) by retaining tokens at regular intervals, effectively halving the density. This process reduces the number of retained tokens in the first two windows from \(2W\) to \(\frac{2W}{2} = W\). 
Subsequently, we add \(W\) new tokens, resulting in a full-precision window size of \(\frac{2W}{2} + W = 2W\). At this point, the densities become \(\text{density}_{W_1} = \frac{1}{2} p\) and \(\text{density}_{W_2} = p\), where \(p\) is the initial density and \(W_i\) denotes the \(i\)-th window. 
By continuously adding new tokens, LogQuant naturally forms a \(\log_2\) sparsity selection within the constrained length. The detailed selection process is described in Algorithm~\hyperref[alg:LogQuant]{1}. Using this approach, the length of retained full-precision tokens fluctuates between \(2W\) and \(3W\), providing a more stable compression ratio compared to KiVi, where the length fluctuates between \(0\) and \(R\), with \(R\) being the length of retained full-precision tokens in KiVi. We illustrate the workflow in {Figure}~\ref{fig:workflow}, which visually represents the KV cache management process, enhancing the understanding of our algorithm's implementation.

\begin{algorithm}
\caption{Log-based Filtering Token Selection Strategy}
\begin{algorithmic}[1]
\State \textbf{Input:} \textit{A} (list of original precision tokens), \textit{a*} (new token), \textit{W} (window length)
\State \textbf{Output:} \textit{A} (updated list of tokens)

\Procedure{AppendToken}{$A$, $a^*$, $W$}
    \If {length(\textit{A}) $<$ $3W$}
        \State \textit{A} \(\leftarrow\) concat(\textit{A}, \textit{a*})
    \Else
        \State \textit{A} \(\leftarrow\) concat(\textit{A}[0:2W:2], \textit{A}[2W:3W])
        \State \textit{A} \(\leftarrow\) concat(\textit{A}, \textit{a*})
    \EndIf
    \State \Return \textit{A}
\EndProcedure
\label{alg:LogQuant}
\end{algorithmic}
\end{algorithm}

\textbf{Implementation.} Popular inference frameworks, such as Hugging Face’s \texttt{transformers} library, have encapsulated KV Cache management into dedicated classes, which simplifies the integration of new methods. To leverage this modular design, we implemented \textbf{LogQuant} as a derived class of the \texttt{Cache} class in the \texttt{transformers} library. This approach ensures seamless compatibility with various quantization backends, including Quanto~\citep{huggingface_optimum_quanto} and HQQ~\citep{badri2023hqq}. For our implementation, we utilized Quanto as the quantization backend, adopting the Key-per-channel strategy. Furthermore, we integrated \textbf{LogQuant} into Hugging Face’s inference pipeline, enhancing its usability for efficient and precise inference workflows.

\section{Experiments}
\label{sec:Experiments}

\subsection{Settings}

\textbf{Models.} We evaluate KiVi and \textit{LogQuant} by 3 popular model families: {Llama3/Llama3.1}~\citep{dubey2024llama3}, {Qwen1.5/Qwen2}~\citep{bai2023qwen, yang2024qwen2}, and {Microsoft Phi3}~\citep{abdin2024phi}. Qwen1.5 and Phi3 are based on Multi-Head Attention, whereas Llama3/3.1 and Qwen2 utilize Group-Query Attention. The quantization group size \(G\) is set to the Hugging Face default value of 64, and the quantized precision is set to INT2. For KiVi, the maximum length of reserved original-precision tokens \(R\) is set to [128, 192, 256]. For LogQuant, the window length \(W\) is limited to \(\lfloor\frac{R}{3}\rfloor\) as it will reserve a maximum of \(3W\) original precision tokens to ensure that the total number of reserved original-precision tokens does not exceed that of KiVi.

\textbf{Datasets.} We selected GSM8K(Grade School Math, ~\citep{GSM8K}) and LongBench~\citep{bai2023longbench} due to their
widespread use in evaluating KV cache quantization,
ensuring our results are comparable to those in the literature. For GSM8K, we test with a 5-shot from the training set for better accuracy and keep the length of the input token between 600 and 1700, the evaluation is based on the exact value of the final answer. For LongBench, we test all 21 datasets among 6 types of tasks and use the LongBench's original pipeline for evaluation. The test dataset details are present in Table~\ref{tab:test_set}.

\subsection{Accuracy and Efficiency Analysis}
\subsubsection{Accuracy Comparison on Different Precision}
\label{sec:diff_pres}
To illustrate the impact of quantized data precision, we evaluate the accuracy loss using Llama3.1-8B-Instruct under both 2-bit and 4-bit quantization for KiVi and LogQuant methods on LongBench. As shown in \autoref{tab:llama-summarized}, both methods achieve performance comparable to the baseline across all tasks with 4-bit quantization. However, 2-bit quantization results in a noticeable drop in accuracy, highlighting the trade-off between memory efficiency and performance. Notably, LogQuant demonstrates better accuracy compared to KiVi under the same conditions.

\begin{table}[ht]
\centering
\caption{Accuracy of Different Precision on Llama3.1-8B. Refer to the \autoref{tab:llama-detailed} for the scores of each specific task. The $\Delta$ shows the difference to baseline.}
\resizebox{1.0\columnwidth}{!}{
\begin{tabular}{lrrrrr}
\hline
\textbf{Category}       & \textbf{KiVi (2-bit)} & \textbf{KiVi (4-bit)} & \textbf{LogQuant (2-bit)} & \textbf{LogQuant (4-bit)} & baseline \\
\hline
\\
Single-Document QA       &  38.89  ($\Delta$ -8.11) &  47.75  ($\Delta$ +0.75) &  41.91  ($\Delta$ -5.09) &  47.73  ($\Delta$ +0.73) &  47.71  \\
Multi-Document QA        &  34.02  ($\Delta$ -4.98) &  39.74  ($\Delta$ +0.74) &  36.08  ($\Delta$ -2.92) &  39.93  ($\Delta$ +0.93) &  39.96  \\
Summarization            &  16.10  ($\Delta$ -1.90) &  17.94  ($\Delta$ -0.06) &  16.62  ($\Delta$ -1.38) &  17.92  ($\Delta$ -0.08) &  18.08  \\
Few-shot Learning        &  52.51  ($\Delta$ -8.49) &  61.34  ($\Delta$ +0.34) &  56.43  ($\Delta$ -4.57) &  61.21  ($\Delta$ +0.21) &  61.22  \\
Synthetic Tasks          &  45.02  ($\Delta$ -21.98) &  67.74  ($\Delta$ +0.74) &  52.51  ($\Delta$ -14.49) &  67.68  ($\Delta$ +0.68) &  67.78  \\
Code Completion          &  43.06  ($\Delta$ -15.94) &  59.53  ($\Delta$ +0.53) &  52.10  ($\Delta$ -6.90) &  59.57  ($\Delta$ +0.57) &  59.78  \\
\hline
\end{tabular}
}
\label{tab:llama-summarized}
\end{table}

\subsubsection{Accuracy Comparison among different Configurations}

As discussed in Section~\ref{sec:diff_pres}, 4-bit quantization incurs only a slight accuracy loss across tasks. Therefore, we focus on 2-bit quantization in the following discussion to highlight LogQuant's performance. To further investigate the accuracy loss resulting from quantization, we compared the following methods:  1) 16-bit baseline, 2) KiVi and 3) LogQuant across different configurations, we define the \textit{compression ratio} as:

\begin{equation}
\begin{aligned}
\frac{\text{Original tensor size}}{\text{Tensor size in compressed format}}\\
\end{aligned}
\end{equation}
where, for a sequence length \(L\) and reserved original precision token length \(R\) in a BF16 model with 2-bit quantization, the \textit{compression ratio} can be expressed as:
\begin{equation}
\begin{aligned}
\frac{16L}{2(L-R)+16R}.\\
\end{aligned}
\end{equation}

\begin{figure}[t]
    \centering
    \includegraphics[width=1\linewidth]{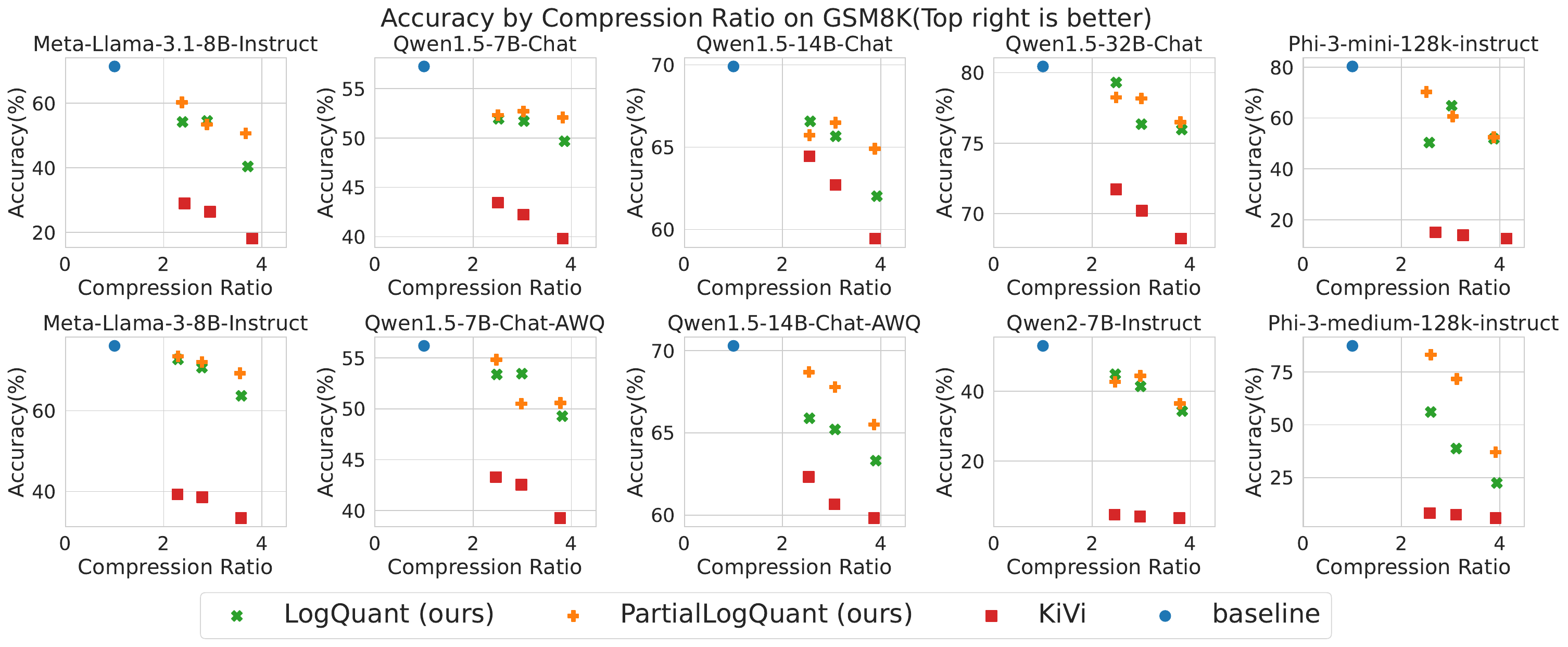}
    \caption{Accuracy(EM) with different compression ratio in GSM8K tasks for different models. }
    \label{fig:compression}
\end{figure}
We tested the three compression ratios using GSM8K across three model families, and the results summarized in Figure~\ref{fig:compression}. Our findings demonstrate that the \textit{LogQuant} method consistently outperforms KiVi across all three models at various compression ratios. The results also indicate that smaller models and small KV states models, such as Phi3-mini (3.8B) and Qwen2-7B (retaining only \(\frac{1}{8}\) of KV heads than Query, while other GQA models typically retain at least \(\frac{1}{4}\).), experience a more significant accuracy loss with 2-bit quantized KV caches. However, our method provides a notable improvement in accuracy for these smaller models.

\subsubsection{Accuracy Comparison among Different Tasks}
To further investigate the accuracy loss across various tasks, we evaluate the seven task groups listed in Table~\ref{tab:test_set} and report the average score for each method in Table~\ref{tab:longbench}. 

In the following, the task groups are abbreviated as follows: Math remains unchanged; Code refers to Code Completion; Few-shot stands for Few-shot Learning; Multi-QA represents Multi-Document QA; Single-QA denotes Single-Document QA; Summ. is short for Summarization; and Synth. stands for Synthetic Tasks. 

We set the reserved length \(R\) to 128, meaning that LogQuant uses only \(3\lfloor\frac{R}{3}\rfloor = 126\) original precision tokens, which is slightly fewer than the 128 tokens reserved by KiVi. As shown in Table~\ref{tab:longbench}, for simpler tasks such as Summarization, quantization has little to no impact on performance compared to the 16-bit baseline. However, for more complex tasks such as Code Completion, Synthetic Tasks, and Math, quantization significantly affects accuracy, with \textit{LogQuant} demonstrating better retention of accuracy than KiVi.

\begin{table}[ht]
\caption{\centering
Task Group Average Score for Different Models with 2-bit KV Cache Quantization. 
(The best result of 2-bit quantization is in bold. Refer to Table~\ref{tab:longbench_all} for the scores of each specific task in LongBench.)}
\centering
\resizebox{1\columnwidth}{!}{
\begin{tabular}{llccccccc}
\hline
\textbf{Model} & \textbf{Method} & \textbf{Math} & \textbf{Code} & \textbf{Few-shot} & \textbf{Multi-QA} & \textbf{Single-QA} & \textbf{Summ.} & \textbf{Synth.} \\ \hline
\multirow{3}{*}{llama-3.1-8B-Instruct} 
& 16-bit Baseline & 71.42 & 59.78 & 61.21 & 39.95 & 47.71 & 18.07 & 67.78 \\
& KiVi           & 18.04 & 43.06 & 52.50 & 34.01 & 38.89 & 16.10 & 45.02 \\
& LogQuant (ours)& \textbf{40.41} & \textbf{52.09} & \textbf{56.42} & \textbf{36.08} & \textbf{41.90} & \textbf{16.62} & \textbf{52.51} \\ \hline

\multirow{3}{*}{Qwen1.5-7B-Chat-AWQ} 
& 16-bit Baseline & 56.18 & 52.46 & 53.88 & 33.05 & 39.26 & 17.11 & 26.50 \\
& KiVi           & 39.27 & 34.79 & 51.32 & 31.08 & 35.80 & 17.16 & 10.00 \\
& LogQuant (ours)& \textbf{49.28} & \textbf{40.68} & \textbf{52.54} & \textbf{32.04} & \textbf{37.22} & \textbf{17.38} & \textbf{13.50} \\ \hline

\multirow{3}{*}{Qwen1.5-14B-Chat-AWQ} 
& 16-bit Baseline & 70.28 & 57.47 & 59.02 & 39.72 & 42.48 & 17.21 & 61.33 \\
& KiVi           & 59.82 & 37.48 & 57.50 & 37.91 & 40.39 & 17.17 & 46.85 \\
& LogQuant (ours)& \textbf{63.31} & \textbf{49.37} & \textbf{58.25} & \textbf{38.01} & \textbf{41.37} & \textbf{17.24} & \textbf{52.17} \\ \hline

\multirow{3}{*}{Qwen2-7B-Instruct} 
& 16-bit Baseline & 52.99 & 58.23 & 61.90 & 33.35 & 44.66 & 16.33 & 43.00 \\
& KiVi           & 3.71  & 35.91 & 35.26 & 12.35 & 20.52 & 9.31  & 11.42 \\
& LogQuant (ours)& \textbf{34.34} & \textbf{48.71} & \textbf{51.23} & \textbf{28.28} & \textbf{34.84} & \textbf{13.13} & \textbf{22.83} \\ \hline

\multirow{3}{*}{Phi-3-mini-128k-instruct} 
& 16-bit Baseline & 80.29 & 55.97 & 52.58 & 33.55 & 42.47 & 17.56 & 48.00 \\
& KiVi           & 12.59 & 33.97 & 36.17 & 18.19 & 19.58 & 9.10  & 4.83  \\
& LogQuant (ours)& \textbf{51.86} & \textbf{40.84} & \textbf{39.36} & \textbf{21.70} & \textbf{23.63} & \textbf{9.89} & \textbf{5.39} \\ \hline
\end{tabular}
}
\label{tab:longbench}
\end{table}

\subsubsection{Efficiency Comparison}
\begin{figure}[t]
    \centering
    \includegraphics[width=0.9\linewidth]{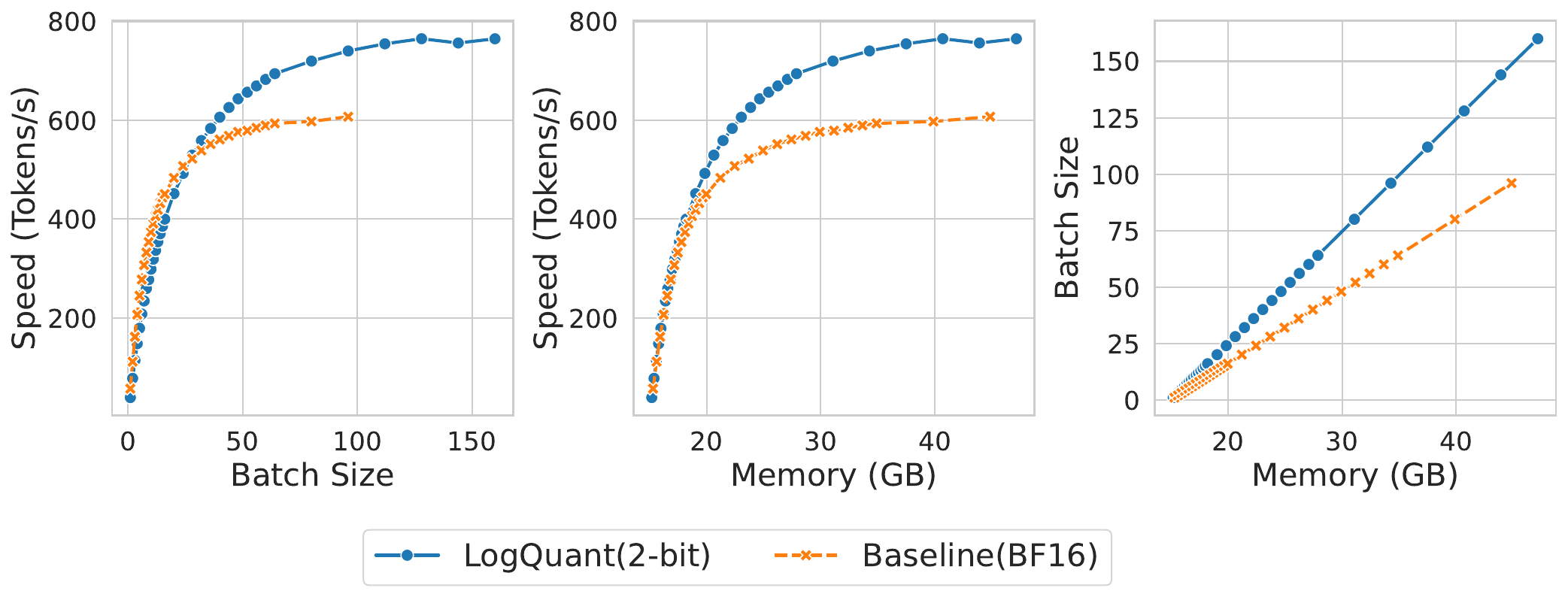}
    \caption{memory usage and throughput comparison between 2bit LogQuant and 16bit baseline under huggingface generation pipeline with llama3.1-8B and H100.}
    \label{fig:speed}
\end{figure}

To evaluate memory and throughput efficiency by a NVIDIA H100 48G MIG with the HuggingFace pipeline, we conducted a benchmark similar to that in \citep{KVbench}, setting an average prompt length of 512 and a maximum output length of 2000. We incrementally increased the batch size while recording peak memory usage and throughput for both \textit{LogQuant} (2-bit with 126 reserved tokens) and the BF16 baseline on the Llama-3.1-8B model, until memory usage reached the 48GB limit. The hardware utilized was a single NVIDIA H100 GPU. As shown in Figure~\ref{fig:speed}, \textit{LogQuant} achieves approximately 25\% higher throughput by supporting a larger batch size. Additionally, it allows for a 60\% increase in batch size within the same memory constraints under the HuggingFace pipeline.

We also observed that, within the HuggingFace pipeline, inference with a quantized cache does not immediately release original KV states, which limits memory compression and efficiency. Furthermore, the dequantization operation impacts throughput. These issues suggest that memory efficiency and speed could be further improved by employing operator fusion, enabling computation on the quantized cache directly with a fused attention operation. We will explore this optimization in future work.

\section{Conclusion and Future Work}
\label{sec:Conclusion}

In this paper, we introduced LogQuant, a novel quantization technique designed to optimize KV Cache management in large language models (LLMs). Our approach leverages a base-2 logarithmic strategy to maintain sparsity while accommodating an increased number of full-precision tokens. Through comprehensive evaluations, we demonstrated that LogQuant consistently outperforms existing methods, such as KiVi, across various model families and compression ratios, particularly benefiting smaller models that typically suffer from accuracy loss due to quantization.

We further explored the efficiency of our implementation within the HuggingFace pipeline, achieving notable improvements in throughput and memory utilization. Additionally, our investigation into accuracy loss across different tasks highlighted LogQuant's superior retention of performance, especially in complex tasks. These findings underscore the potential of LogQuant to enhance LLM inference in resource-constrained environments.

Future work will focus on refining our quantization approach and investigating further optimizations, such as operator fusion, to maximize efficiency and performance in LLM applications.

\bibliographystyle{iclr2025_conference}
\bibliography{reference}
%\bibliographystyle{iclr2025_conference}
%\bibliography{reference}

\newpage
% \newpage
\appendix

\section{Background \& Related Work: KV Cache Compression}
\label{sec:RelatedWork}
The attention mechanism relies on three key components: the Query (Q), Key (K), and Value (V) vectors. For each token, LLM computes a $d$-dimensional Q vector and compares it against all stored $N \times d$ K vectors, where $N$ is the length of the sequence processed. The result of this comparison is used to weigh the corresponding V vectors, producing the final output. Mathematically, the attention operation is defined as:

\begin{equation}
\begin{aligned}
\text{Attention}(Q, K, V) = \text{Softmax}\left( \frac{Q K^\top}{\sqrt{d}} \right) V\\
\end{aligned}
\end{equation}

LLM inference is generally divided into two phases: a prefill phase for processing input tokens and a decoding phase for generating new tokens. In decoding, each token generation reloads the entire KV Cache from previous tokens, causing time and memory inefficiencies.

KV cache compression methods fall into two categories: 'training-free' methods (using eviction and quantization without model retraining) and 'training-required' methods (designing more efficient attention structures). Our approach focuses on enhancing training-free methods for broader applicability. Eviction selectively discards less important tokens, while quantization lowers the precision of key and value states to save memory. However, both methods risk significant information loss at high compression rates—especially 2-bit quantization, which can greatly reduce accuracy.

\subsection{KV Cache Eviction}
Eviction methods aim to reduce KV cache memory usage in Large Language Models (LLMs) by discarding less important tokens. The early work H2O~\citep{zhang2024h2o} selects "heavy hitter" tokens based on cumulative attention scores, though this risks evicting tokens that may become important later. Keyformer~\citep{adnan2024keyformer} improves on H2O by combining "Key Attention" with a "window attention" mechanism, retaining both historically significant and recent tokens for better accuracy. MiniCache~\citep{liu2024minicache} reduces memory by reusing Key and Value states across layers. This method assumes that some key and value representations are redundant across model layers and can be shared. InfLLM~\citep{xiao2024infllm} addresses very long contexts by dividing them into blocks and retaining 'representative tokens' for block eviction decisions.

\subsection{KV Cache Quantization}
Quantization reduces storage and boosts computational speed by using fewer bits to represent values. Earlier works, like {AWQ}~\citep{lin2023awq} and {Qserve}~\citep{lin2024qserve}, applied 4-bit quantization to the KV cache with minimal accuracy loss. Recent methods aim to compress the KV cache further while preserving accuracy. {QAQ}~\citep{dong2024qaq} dynamically adjusts the precision of the in-GPU quantized cache by offloading all original-precision KV data to CPU memory. {GEAR}~\citep{kang2024gear} improves accuracy by storing the quantization error of the KV cache as a sparse matrix with low-rank decomposition. {KiVi}~\citep{liu2024kivi} introduces a 2-bit quantization by retaining a recent window of full-precision tokens, balancing memory efficiency and accuracy.

\subsection{training-required approaches}
An early memory-reducing attention design is Multi-Query Attention ({MQA,}~\citep{shazeer2019fast}), where all query heads share a single pair of key and value heads. While this reduces memory, it significantly impacts accuracy. Grouped-Query Attention ({GQA,}~\citep{ainslie2023gqa}) addresses this by grouping query heads, with each group sharing the same key and value heads, preserving the generalization ability of multi-head attention while reducing KV cache size. {Deepseek V2}~\citep{deepseekai2024deepseekv2strongeconomicalefficient} introduces Multi-Head Latent Attention (MLA), which compresses key and value states using LoRA-based projections. To prevent disruption of position embeddings from LoRA compression, specific channels are reserved for position information only, excluding them from LoRA compression.

\newpage

\renewcommand{\thetable}{B\arabic{table}}
\section{Overview of Test Datasets}

\begin{table}[h]
\caption{\centering{Overview of all test datasets.} {‘Avg len’ (average length) is computed using the number of words for the English (code) datasets and the number of characters for the Chinese datasets. ‘Accuracy (CLS)’ refers to classification accuracy, while ‘Accuracy (EM)’ refers to exact match accuracy}}
\centering
\resizebox{1\columnwidth}{!}{
\begin{tabular}{llrllr}
\hline
\textbf{Task Group} & \textbf{Dataset}       & \textbf{Avg len} & \textbf{Metric}   & \textbf{Language} & \textbf{\#data} \\ \hline
\\
\multirow{1}{*}{\textbf{Math}}                                                   
& GSM8K                  & 240             & Accuracy (EM)      & English           & 1319            \\ \hline
\multirow{4}{*}{\textbf{Single-Document QA}}                                                    
&NarrativeQA          & 18,409          & F1                & English           & 200            \\ 
&Qasper                 & 3,619           & F1                & English           & 200            \\ 
&MultiFieldQA-en        & 4,559           & F1                & English           & 150            \\ 
&MultiFieldQA-zh        & 6,701           & F1                & Chinese           & 200            \\ \hline
\multirow{4}{*}{\textbf{Multi-Document QA}}                                                    
&HotpotQA               & 9,151           & F1                & English           & 200            \\ 
&2WikiMultihopQA        & 4,887           & F1                & English           & 200            \\ 
&MuSiQue                & 11,214          & F1                & English           & 200            \\ 
&DuReader               & 15,768          & Rouge-L           & Chinese           & 200            \\ \hline
\multirow{4}{*}{\textbf{Summarization}}                                                         
&GovReport              & 8,734           & Rouge-L           & English           & 200            \\ 
&QMSum                  & 10,614          & Rouge-L           & English           & 200            \\ 
&MultiNews              & 2,113           & Rouge-L           & English           & 200            \\ 
&VCSUM                  & 15,380          & Rouge-L           & Chinese           & 200            \\ \hline
\multirow{4}{*}{\textbf{Few-shot Learning}}                                                     
&TREC                   & 5,177           & Accuracy (CLS)    & English           & 200            \\ 
&TriviaQA               & 8,209           & F1                & English           & 200            \\ 
&SAMSum                 & 6,258           & Rouge-L           & English           & 200            \\
&LSHT                   & 22,337          & Accuracy (CLS)    & Chinese           & 200            \\ \hline
\multirow{3}{*}{\textbf{Synthetic Task}}                                                        
&PassageCount           & 11,141          & Accuracy (EM)     & English           & 200            \\
&PassageRetrieval-en     & 9,289           & Accuracy (EM)     & English           & 200            \\
&PassageRetrieval-zh     & 6,745           & Accuracy (EM)     & Chinese           & 200            \\ \hline
\multirow{2}{*}{\textbf{Code Completion}}                                                       
&LCC                    & 1,235           & Edit Sim          & Python/C\#/Java   & 500            \\
&RepoBench-P            & 4,206           & Edit Sim          & Python/Java       & 500            \\ \hline
\end{tabular}
}
\label{tab:test_set}
\end{table}

\renewcommand{\thetable}{C\arabic{table}}
\section{Meta Data of Precision Comparison}

\begin{table}[h]
\centering
\caption{Comparison on Llama3.1-8B-Instruct of different quantization precisions}
\resizebox{1\columnwidth}{!}{
\begin{tabular}{lrrrrr}
\hline
\textbf{Dataset}              & \textbf{KiVi (2-bit)} & \textbf{KiVi (4-bit)} & \textbf{LogQuant (2-bit)} & \textbf{LogQuant (4-bit)} & \textbf{Baseline} \\
\hline
2wikimqa             & 39.52 & 44.79 & 40.69 & 45.18 & 45.06 \\
dureader             & 22.20 & 27.75 & 22.59 & 27.99 & 28.48 \\
gov\_report          & 18.60 & 19.86 & 18.78 & 20.09 & 20.41 \\
hotpotqa             & 48.83 & 55.78 & 52.43 & 55.85 & 55.90 \\
lcc                  & 47.09 & 63.44 & 57.52 & 62.85 & 62.99 \\
lsht                 & 31.42 & 45.00 & 33.75 & 45.00 & 45.00 \\
multi\_news          & 15.07 & 15.65 & 15.11 & 15.64 & 15.89 \\
multifieldqa\_en     & 42.51 & 55.10 & 45.98 & 54.63 & 54.91 \\
multifieldqa\_zh     & 50.12 & 62.77 & 55.51 & 63.27 & 62.72 \\
musique              & 25.52 & 30.65 & 28.62 & 30.70 & 30.39 \\
narrativeqa          & 26.44 & 27.91 & 27.93 & 28.28 & 28.19 \\
passage\_count       & 5.67  & 6.31  & 5.63  & 6.15  & 6.31  \\
passage\_retrieval\_en & 83.17 & 99.50 & 92.25 & 99.50 & 99.50 \\
passage\_retrieval\_zh & 46.23 & 97.42 & 59.65 & 97.38 & 97.54 \\
qasper               & 36.50 & 45.20 & 38.21 & 44.74 & 45.03 \\
qmsum                & 17.41 & 19.07 & 18.19 & 18.92 & 19.15 \\
repobench-p          & 39.03 & 55.61 & 46.67 & 56.28 & 56.57 \\
samsum               & 23.88 & 36.12 & 33.33 & 35.45 & 35.72 \\
trec                 & 65.00 & 72.50 & 67.00 & 72.50 & 72.50 \\
triviaqa             & 89.72 & 91.73 & 91.63 & 91.89 & 91.64 \\
vcsum                & 13.33 & 17.17 & 14.41 & 17.04 & 16.85 \\
\hline
\end{tabular}
}
\label{tab:llama-detailed}
\end{table}

\renewcommand{\thetable}{D\arabic{table}}
\section{Meta Data of LongBench Results}

\begin{center}
\begin{longtable}{l|rrr}
    \caption{LongBench score of each dataset} \label{tab:longbench_all} \\
    
    \hline  
    \textbf{precision}        & \textbf{16-bit} & \multicolumn{2}{c}{\textbf{2-bit}} \\                
    \textbf{Task Group}        & \textbf{Baseline} & \textbf{KiVi} & \makecell[r]{\textbf{LogQuant} \\ \textbf{(ours)}} \\ \hline
    \endfirsthead
    
    \multicolumn{4}{c}%
    {{\bfseries \tablename\ \thetable{} -- continued from previous page}} \\
    \hline             
    \textbf{Task Group}        & \textbf{Baseline} & \textbf{KiVi} & \makecell[r]{\textbf{LogQuant} \\ \textbf{(ours)}} \\ \hline 
    \endhead
    
    \multicolumn{2}{c}{{Continued on next page}} \\ 
    \endfoot
    
    \hline
    \endlastfoot

    \multicolumn{4}{c}{\textbf{llama-3-8B-Instruct}}\\
    \hline
    2WikiMultihopQA & 37.24 & 31.72 & \textbf{35.08} \\
    DuReader & 16.73 & 12.45 & \textbf{15.5} \\
    GovReport & 17.8 & 12.8 & \textbf{15.63} \\
    HotpotQA & 46.1 & 43.87 & \textbf{44.96} \\
    LCC & 56.85 & 31.73 & \textbf{41.75} \\
    LSHT & 25.25 & 21.5 & \textbf{21.75} \\
    MultiFieldQA-en & 44.44 & 38.68 & \textbf{41.04} \\
    MultiFieldQA-zh & 56.3 & 43.96 & \textbf{48.44} \\
    MultiNews & 16.59 & 15.76 & \textbf{16.06} \\
    MuSiQue & 21.44 & 19.56 & \textbf{20.59} \\
    NarrativeQA & 22.07 & 19.82 & \textbf{21.56} \\
    PassageCount & 6.5 & \textbf{5.5} & 4.0 \\
    PassageRetrieval-en & 66.0 & 53.0 & \textbf{58.5} \\
    PassageRetrieval-zh & 91.0 & 33.45 & \textbf{72.0} \\
    Qasper & 43.69 & 33.9 & \textbf{39.46} \\
    QMSum & 17.49 & 17.01 & \textbf{17.37} \\
    RepoBench-P & 51.32 & 31.99 & \textbf{40.1} \\
    SAMSum & 33.22 & 22.44 & \textbf{32.66} \\
    TREC & 74.0 & 72.5 & \textbf{73.0} \\
    TriviaQA & 90.48 & 87.65 & \textbf{89.36} \\
    VCSUM & 0.16 & 0.17 & \textbf{0.25} \\
    \hline
    \multicolumn{4}{c}{\textbf{llama-3.1-8B-Instruct}}\\
    \hline
    2WikiMultihopQA & 45.06 & 39.52 & \textbf{40.69} \\
    DuReader & 28.48 & 22.2 & \textbf{22.59} \\
    GovReport & 20.41 & 18.6 & \textbf{18.78} \\
    HotpotQA & 55.9 & 48.83 & \textbf{52.43} \\
    LCC & 62.99 & 47.09 & \textbf{57.52} \\
    LSHT & 45.0 & 31.42 & \textbf{33.75} \\
    MultiFieldQA-en & 54.91 & 42.51 & \textbf{45.98} \\
    MultiFieldQA-zh & 62.72 & 50.12 & \textbf{55.51} \\
    MultiNews & 15.89 & 15.07 & \textbf{15.11} \\
    MuSiQue & 30.39 & 25.52 & \textbf{28.62} \\
    NarrativeQA & 28.19 & 26.44 & \textbf{27.93} \\
    PassageCount & 6.31 & \textbf{5.67} & 5.63 \\
    PassageRetrieval-en & 99.5 & 83.17 & \textbf{92.25} \\
    PassageRetrieval-zh & 97.54 & 46.23 & \textbf{59.65} \\
    Qasper & 45.03 & 36.5 & \textbf{38.21} \\
    QMSum & 19.15 & 17.41 & \textbf{18.19} \\
    RepoBench-P & 56.57 & 39.03 & \textbf{46.67} \\
    SAMSum & 35.72 & 23.88 & \textbf{33.33} \\
    TREC & 72.5 & 65.0 & \textbf{67.0} \\
    TriviaQA & 91.64 & 89.72 & \textbf{91.63} \\
    VCSUM & 16.85 & 13.33 & \textbf{14.41} \\
    \hline
    \multicolumn{4}{c}{\textbf{Phi-3-mini-128k-instruct}}\\
    \hline
    2WikiMultihopQA & 35.78 & 19.12 & \textbf{24.61} \\
    DuReader & 22.75 & \textbf{10.38} & 9.26 \\
    GovReport & 18.7 & 8.83 & \textbf{9.47} \\
    HotpotQA & 50.44 & 31.33 & \textbf{37.48} \\
    LCC & 57.44 & 39.85 & \textbf{47.53} \\
    LSHT & 27.25 & 14.25 & \textbf{13.75} \\
    MultiFieldQA-en & 54.9 & 29.04 & \textbf{34.91} \\
    MultiFieldQA-zh & 52.09 & 8.16 & \textbf{12.32} \\
    MultiNews & 15.52 & 12.72 & \textbf{13.33} \\
    MuSiQue & 25.23 & 11.92 & \textbf{15.46} \\
    NarrativeQA & 23.28 & 15.34 & \textbf{17.37} \\
    PassageCount & 3.0 & 2.25 & \textbf{4.5} \\
    PassageRetrieval-en & 82.5 & 11.0 & \textbf{9.68} \\
    PassageRetrieval-zh & 58.5 & 1.25 & \textbf{2.0} \\
    Qasper & 39.6 & 25.78 & \textbf{29.91} \\
    QMSum & 17.97 & 5.88 & \textbf{7.04} \\
    RepoBench-P & 54.49 & 28.09 & \textbf{34.16} \\
    SAMSum & 30.62 & 9.23 & \textbf{13.03} \\
    TREC & 66.0 & 59.5 & \textbf{62.5} \\
    TriviaQA & 86.43 & 61.72 & \textbf{68.15} \\
    VCSUM & 18.04 & 8.97 & \textbf{9.74} \\
    \hline
    \multicolumn{4}{c}{\textbf{Qwen1.5-14B-Chat-AWQ}}\\
    \hline
    2WikiMultihopQA & 44.81 & 44.35 & \textbf{44.39} \\
    DuReader & 26.02 & 23.34 & \textbf{23.28} \\
    GovReport & 16.31 & 16.23 & \textbf{16.25} \\
    HotpotQA & 55.67 & 53.69 & \textbf{53.9} \\
    LCC & 56.69 & 36.94 & \textbf{50.95} \\
    LSHT & 37.0 & 32.5 & \textbf{34.5} \\
    MultiFieldQA-en & 48.36 & 44.75 & \textbf{45.68} \\
    MultiFieldQA-zh & 60.35 & 58.54 & \textbf{59.43} \\
    MultiNews & 14.95 & \textbf{15.01} & 14.94 \\
    MuSiQue & 32.38 & 30.25 & \textbf{30.45} \\
    NarrativeQA & 22.26 & 21.73 & \textbf{22.83} \\
    PassageCount & 1.0 & \textbf{2.55} & 2.0 \\
    PassageRetrieval-en & 94.5 & 71.0 & \textbf{80.0} \\
    PassageRetrieval-zh & 88.5 & 67.0 & \textbf{74.5} \\
    Qasper & 38.93 & 36.56 & \textbf{37.54} \\
    QMSum & 18.16 & 18.03 & \textbf{18.13} \\
    RepoBench-P & 58.25 & 38.03 & \textbf{47.79} \\
    SAMSum & 32.95 & 32.69 & \textbf{33.34} \\
    TREC & 77.5 & 76.5 & \textbf{77.5} \\
    TriviaQA & 88.63 & \textbf{88.32} & 87.66 \\
    VCSUM & 19.41 & 19.42 & \textbf{19.65} \\
    \hline
    \multicolumn{4}{c}{\textbf{Qwen1.5-7B-Chat}}\\
    \hline
    2WikiMultihopQA & 32.8 & 31.83 & \textbf{32.14} \\
    DuReader & 25.96 & 22.64 & \textbf{24.06} \\
    GovReport & 16.66 & 15.57 & \textbf{15.84} \\
    HotpotQA & 48.11 & 47.37 & \textbf{48.91} \\
    LCC & 58.17 & 45.87 & \textbf{53.77} \\
    LSHT & 28.0 & 24.0 & \textbf{24.5} \\
    MultiFieldQA-en & 47.14 & 42.26 & \textbf{43.72} \\
    MultiFieldQA-zh & 53.4 & 50.18 & \textbf{51.68} \\
    MultiNews & 15.02 & \textbf{15.0} & 14.92 \\
    MuSiQue & 26.74 & 25.88 & \textbf{27.09} \\
    NarrativeQA & 20.06 & 19.02 & \textbf{20.06} \\
    PassageCount & 1.0 & \textbf{0.5} & 0.0 \\
    PassageRetrieval-en & 40.5 & 20.0 & \textbf{24.0} \\
    PassageRetrieval-zh & 59.0 & 18.25 & \textbf{29.0} \\
    Qasper & 39.84 & 37.19 & \textbf{37.28} \\
    QMSum & 18.25 & 17.59 & \textbf{18.18} \\
    RepoBench-P & 45.46 & 26.33 & \textbf{30.76} \\
    SAMSum & 33.01 & 29.7 & \textbf{33.31} \\
    TREC & 70.5 & \textbf{69.5} & 67.5 \\
    TriviaQA & 86.76 & 86.51 & \textbf{87.37} \\
    VCSUM & 17.98 & 19.15 & \textbf{19.34} \\
    \hline
    \multicolumn{4}{c}{\textbf{Qwen1.5-7B-Chat-AWQ}}\\
    \hline
    2WikiMultihopQA & 32.43 & 30.82 & \textbf{33.46} \\
    DuReader & 25.84 & 23.1 & \textbf{24.36} \\
    GovReport & 16.98 & 16.31 & \textbf{16.65} \\
    HotpotQA & 47.77 & \textbf{47.17} & 46.0 \\
    LCC & 57.98 & 44.56 & \textbf{52.33} \\
    LSHT & 29.0 & 25.5 & \textbf{27.0} \\
    MultiFieldQA-en & 46.72 & 42.87 & \textbf{45.85} \\
    MultiFieldQA-zh & 50.97 & 45.51 & \textbf{46.73} \\
    MultiNews & 14.97 & 15.04 & \textbf{15.16} \\
    MuSiQue & 26.18 & 23.23 & \textbf{24.36} \\
    NarrativeQA & 20.93 & 19.58 & \textbf{20.14} \\
    PassageCount & 0.5 & \textbf{0.0} & 0.0 \\
    PassageRetrieval-en & 30.5 & 16.0 & \textbf{18.5} \\
    PassageRetrieval-zh & 48.5 & 14.0 & \textbf{22.0} \\
    Qasper & 38.45 & 35.27 & \textbf{36.16} \\
    QMSum & 17.85 & 17.34 & \textbf{17.77} \\
    RepoBench-P & 46.95 & 25.02 & \textbf{29.03} \\
    SAMSum & 31.98 & 28.3 & \textbf{32.06} \\
    TREC & 67.0 & \textbf{65.0} & 63.5 \\
    TriviaQA & 87.56 & 86.48 & \textbf{87.61} \\
    VCSUM & 18.66 & 19.95 & \textbf{19.96} \\
    \hline
    \multicolumn{4}{c}{\textbf{Qwen2-7B-Instruct}}\\
    \hline
    2WikiMultihopQA & 44.15 & 11.33 & \textbf{40.12} \\
    DuReader & 19.22 & 13.08 & \textbf{15.01} \\
    GovReport & 18.09 & 10.82 & \textbf{16.07} \\
    HotpotQA & 44.3 & 17.39 & \textbf{39.92} \\
    LCC & 57.72 & 36.63 & \textbf{51.46} \\
    LSHT & 44.0 & 23.0 & \textbf{26.25} \\
    MultiFieldQA-en & 46.89 & 21.97 & \textbf{36.42} \\
    MultiFieldQA-zh & 61.48 & 33.67 & \textbf{47.57} \\
    MultiNews & 15.58 & 8.53 & \textbf{13.6} \\
    MuSiQue & 25.71 & 7.58 & \textbf{18.07} \\
    NarrativeQA & 24.43 & 5.29 & \textbf{18.43} \\
    PassageCount & 5.0 & 5.5 & \textbf{5.5} \\
    PassageRetrieval-en & 69.0 & 19.25 & \textbf{33.5} \\
    PassageRetrieval-zh & 55.0 & 9.5 & \textbf{29.5} \\
    Qasper & 45.82 & 21.16 & \textbf{36.94} \\
    QMSum & 17.92 & 9.08 & \textbf{12.25} \\
    RepoBench-P & 58.74 & 35.18 & \textbf{45.95} \\
    SAMSum & 35.94 & 18.23 & \textbf{28.03} \\
    TREC & 78.0 & 58.25 & \textbf{68.0} \\
    TriviaQA & 89.66 & 41.56 & \textbf{82.63} \\
    VCSUM & 13.74 & 8.82 & \textbf{10.58} \\
    \hline
\end{longtable}
\end{center}

\end{document}